\documentclass[11pt]{article}
\usepackage[margin=1.1in]{geometry}
\usepackage{amsmath,amssymb,amsthm,mathrsfs}
\usepackage{booktabs}
\usepackage{graphicx}
\usepackage[numbers,sort&compress]{natbib}
\usepackage[colorlinks=true,linkcolor=blue,citecolor=blue,urlcolor=blue]{hyperref}

\newtheorem{theorem}{Theorem}[section]
\newtheorem{proposition}[theorem]{Proposition}
\newtheorem{lemma}[theorem]{Lemma}
\newtheorem{corollary}[theorem]{Corollary}
\newtheorem{definition}[theorem]{Definition}

\theoremstyle{remark}
\newtheorem{remark}[theorem]{Remark}

\newcommand{\Hs}{\mathscr{H}}

\newcommand{\E}{\mathbb{E}}
\newcommand{\Var}{\operatorname{Var}}
\newcommand{\Meff}{M_{\mathrm{eff}}}

\title{\bf On Fibonacci Ensembles\\[2mm]
\large An Alternative Approach to Ensemble Learning Inspired by the\\
Timeless Architecture of the Golden Ratio}

\author{Ernest Fokou\'e\\
\small School of Mathematics and Statistics, Rochester Institute of Technology\\
\small \texttt{epfeqa@rit.edu}}

\date{}

\begin{document}
\maketitle

\begin{abstract}
\noindent
Nature offers the Fibonacci sequence as a glimpse of recursive order, and it is
natural to ask whether the same architecture might organize the way many
imperfect learners are combined into one. This paper takes that question
seriously and answers it precisely. We introduce \emph{Fibonacci ensembles},
in which base learners drawn from an ordered dictionary receive weights
proportional to the Fibonacci numbers, and we characterize exactly what such
weighting can and cannot achieve.

Our central structural result is that normalized Fibonacci weights possess a
\emph{fixed effective ensemble size},
$\Meff \to \varphi^{3} = 2+\sqrt5 \approx 4.236$, independent of the number $M$
of available learners. Fibonacci weighting consults, in effect, slightly more
than four learners no matter how many are supplied. It therefore cannot deliver
the $O(1/M)$ variance decay that justifies bagging, and uniform weighting
strictly dominates it in variance for every $M > \varphi^3$. Any advantage of
structured weighting must instead come from bias reduction on a dictionary whose
index carries genuine meaning. We make this precise through an exact
bias--variance crossover condition for the geometric family, and we show that
normalized Fibonacci weights are numerically indistinguishable from geometric
weights of ratio $\varphi$ once $M \ge 20$.

We further develop second-order recursive ensemble flows, establish their
stability region via the Schur--Cohn conditions, observe that the classical
Fibonacci recursion lies outside it, and construct a damped flow that preserves
the golden mode ratio $\varphi^{2}$ while remaining $L^2$-stable. Controlled
experiments with random Fourier feature and polynomial dictionaries show that the
weighting law moves integrated squared error by more than an order of magnitude,
that the risk-optimal geometric ratio ranges from $0.20$ to $2.20$ depending on
dictionary, target and $M$, and that $\varphi$ is nowhere the minimizer. A
permutation study documents severe sensitivity to the ordering of the
dictionary.

Measured against classical ordered shrinkage, the picture sharpens further. The
geometric family is very nearly optimal \emph{within} the class of convex
aggregation rules, coming within $12\%$ of the best simplex-constrained
estimator; but that entire class is between one and two orders of magnitude worse
than Pinsker or blockwise Stein shrinkage, because the constraint $\sum_m w_m=1$
is itself the binding limitation. On three real series---Nile river flows,
sunspot activity, and Pacific sea-surface temperature---structured weighting
improves on uniform averaging where the dictionary ordering is informative and
harms it where it is not.

The conclusion we draw is more modest, and we think more useful, than a claim of
golden-ratio optimality. The weighting law is a consequential and largely
undesigned component of ensemble learning; Fibonacci weighting is a tuning-free
and interpretable default within it; and estimating the law from data is the
substantive problem that remains open. The golden ratio does not disappear from
the theory---it reappears, as the exact ceiling on how much of a dictionary such
a scheme can ever use.
\end{abstract}


\tableofcontents

\section{Introduction: Learning in the Light of an Eternal Sequence}

The Fibonacci sequence is among humanity's earliest encounters with the silent
mathematics of growth. In the spirals of shells, the phyllotaxis of plants, and
the harmonics of the golden ratio, its recurrence suggests that stability and
creativity can spring from recursive simplicity
\citep{Koshy2001Fibonacci, Livio2002GoldenRatio}.

Ensemble learning is itself an act of aggregation: a formal attempt to let many
imperfect learners speak with one voice. Bagging \citep{Breiman1996Bagging},
boosting \citep{FreundSchapire1997, Friedman2001GBM}, and random forests
\citep{Breiman2001RandomForests} have shown that combining weak or moderately
strong predictors yields substantial gains; see also
\citet{HastieTibshiraniFriedman2009ESL} and \citet{Zhou2012Ensemble}. Yet the
aggregation structure is rarely designed. It is inherited: uniform for bagging,
implicit in the loss for boosting, estimated as a nuisance parameter for
stacking.

This paper asks what happens when the aggregation rule is treated as a
first-class object, and it anchors that question in a specific and unusually
tractable choice---weights proportional to the Fibonacci numbers. The choice is
attractive for reasons partly aesthetic and partly mathematical. Fibonacci is
the canonical positive integer sequence obeying a second-order recursion, it
requires no tuning, and its growth rate is the slowest of the classical
geometric laws. It is a reasonable hunch, and the purpose of what follows is to
determine, carefully and quantitatively, whether it survives.

The answer has two halves. Structurally, Fibonacci weighting turns out to be far
more constrained than its geometric picture suggests: it concentrates on a
bounded number of learners, and the bound is $\varphi^3 \approx 4.236$. This
single fact reorganizes the entire theory, because it removes variance reduction
from the list of things such a scheme can accomplish and places the whole
argument on bias. Empirically, once one recognizes that Fibonacci weighting is
geometric weighting at ratio $\varphi$, the natural question is whether $\varphi$
is a preferred ratio; systematic sweeps say that it is not, and that the optimal
ratio depends on the problem in ways that are interpretable through the
bias--variance condition we derive.

We regard neither half as a disappointment. The first is a clean and previously
unrecorded piece of mathematics about a classical sequence. The second closes a
question that someone would otherwise have had to open.

\subsection{Contributions}

\begin{enumerate}
\item \textbf{Fixed effective size.} Normalized Fibonacci weights satisfy
$\sum_m \alpha_m^2 \to \varphi^{-3}$ and hence $\Meff \to \varphi^3 = 2+\sqrt5$,
independently of $M$ (Theorem~\ref{thm:effsize}). Uniform weighting strictly
dominates in variance for all $M > \varphi^3$ (Corollary~\ref{cor:novar}).

\item \textbf{Exact crossover.} We give a necessary and sufficient condition for
a geometric law to improve on uniform weighting in integrated risk
(Theorem~\ref{thm:crossover}), showing that such laws are bias-reduction devices
that pay a variance premium growing with $M$.

\item \textbf{Equivalence to geometric weighting.} Fibonacci and geometric-$\varphi$
weights agree to $1.6\times10^{-5}$ at $M=20$ and to machine precision for
$M\ge40$ (Section~\ref{sec:exp}), so any Fibonacci-specific claim is a claim
about the number $\varphi$.

\item \textbf{Stability of second-order flows.} We establish the stability region
via the Schur--Cohn conditions (Proposition~\ref{prop:stab}), note that
$(\beta,\gamma)=(1,1)$ lies outside it, and construct a damped flow retaining the
golden mode ratio $\varphi^2$ while remaining $L^2$-stable
(Proposition~\ref{prop:damped}).

\item \textbf{Empirical characterization.} Sweeps over the geometric family show
ISE varying by more than an order of magnitude, optimal ratios from $0.20$ to
$2.20$, and $\varphi$ nowhere optimal; permutation studies show $2.5\times$ to
$15\times$ swings under reordering.

\item \textbf{Comparison with classical shrinkage.} Against Pinsker, blockwise
Stein and oracle benchmarks (Section~\ref{sec:classical}), the geometric family
is within $12\%$ of the best possible simplex-constrained estimator, while the
simplex constraint itself costs a factor of $6$ to $146$. The constraint, not the
law, is the binding limitation.

\item \textbf{Real data.} On Nile river flows, sunspot activity and Pacific
sea-surface temperature (Section~\ref{sec:real}), structured weighting helps
where the ordering is informative and hurts where it is not, and the
cross-validated ratio exceeds $\varphi$ on all three series.
\end{enumerate}

\subsection{Related Work}

Combining predictors is classical. Bagging \citep{Breiman1996Bagging} reduces
variance by averaging unstable learners; random forests
\citep{Breiman2001RandomForests} add randomized feature subsampling; gradient
boosting \citep{Friedman2001GBM} builds an additive model by stagewise
functional gradient descent. Stacked generalization
\citep{Wolpert1992Stacking, Breiman1996StackedReg} trains a meta-learner over
base predictions, and the super learner \citep{vanderLaan2007SuperLearner}
provides an asymptotically optimal convex combination via cross-validated risk
minimization. Bayesian model averaging \citep{Hoeting1999BMA} derives weights
from posterior model probabilities.

Closest to the present concern is the literature on weighted aggregation over an
\emph{ordered} dictionary, which we wish to place squarely in view. Exponential
and geometric weighting over a model index is developed by
\citet{Catoni2004SaintFlour}, \citet{Yang2001Mixing},
\citet{LeungBarron2006}, and \citet{DalalyanTsybakov2008}, the last three
establishing oracle inequalities for essentially the problem considered here.
Optimal weight sequences over an ordered basis are classical in nonparametric
estimation: Pinsker weights \citep{Pinsker1980}, blockwise Stein shrinkage, and
the ordered-linear-smoother theory of \citet{Kneip1994Ordered} and
\citet{CavalierTsybakov2002} all address how influence should decay along an
index, with genuine optimality guarantees. In model averaging,
\citet{Hansen2007MMA} and \citet{HansenRacine2012JMA} derive data-driven weights
with asymptotic optimality.

Against this background, the contribution of Fibonacci weighting is not that it
is the first structured law---it plainly is not---but that it is a
\emph{closed-form, tuning-free} member of a family whose optimum is normally
estimated, and that its extreme concentration makes the limits of the whole
family unusually easy to see. The companion manuscripts
\citep{FokoueII, FokoueIII} develop, respectively, the general weighting theory
of which this is an instance, and the dynamical theory of the recursive flows
introduced in Section~\ref{sec:flows}.

\section{Fibonacci Weighting and Its Geometry}

Let $h_1,\dots,h_M$ be base learners in a real Hilbert space
$\Hs \subseteq L^2(P_X)$ with $\langle f,g\rangle = \E[f(X)g(X)]$. Throughout,
the index $m$ is assumed to carry meaning---typically increasing complexity,
frequency, or degrees of freedom. Section~\ref{sec:perm} shows that this
assumption is not cosmetic.

\begin{definition}[Fibonacci weighting]
With $F_1=F_2=1$ and $F_m = F_{m-1}+F_{m-2}$, set
\[
\alpha_m = \frac{F_m}{\sum_{j=1}^M F_j},
\qquad
\widehat f_{\mathrm{Fib}}(x) = \sum_{m=1}^M \alpha_m h_m(x).
\]
\end{definition}

\begin{definition}[Effective ensemble size]\label{def:eff}
For weights $w$ on the simplex, $\Meff(w) = \big(\sum_m w_m^2\big)^{-1}$.
\end{definition}

By Cauchy--Schwarz, $\Meff(w)\le M$ with equality iff $w$ is uniform. This
quantity governs what a weighting law can accomplish.

\subsection{Golden Asymptotics}

\begin{lemma}[Golden asymptotic equidistribution]\label{lem:golden}
For fixed $k\ge0$, as $M\to\infty$, $\alpha_{M-k} \to \varphi^{-(k+2)}$, where
$\varphi = (1+\sqrt5)/2$.
\end{lemma}

\begin{proof}
Since $\sum_{j=1}^M F_j = F_{M+2}-1$ and
$F_m = \varphi^m/\sqrt5 + O(\varphi^{-m})$,
\[
\alpha_{M-k} = \frac{F_{M-k}}{F_{M+2}-1}
\longrightarrow \frac{\varphi^{M-k}}{\varphi^{M+2}} = \varphi^{-(k+2)} .
\]
Consistency: $\sum_{k\ge0}\varphi^{-(k+2)} = \varphi^{-2}(1-\varphi^{-1})^{-1}=1$.
\end{proof}

The weight profile therefore converges to a fixed geometric shape anchored at the
\emph{end} of the dictionary, with $\alpha_M \to \varphi^{-2}\approx 0.382$ and
$\alpha_{M-1}\to\varphi^{-3}\approx 0.236$. The final two learners jointly carry
about $62\%$ of the mass regardless of $M$.

\subsection{The Effective Size Theorem}

\begin{theorem}[Fixed effective ensemble size]\label{thm:effsize}
For normalized Fibonacci weights,
\[
\sum_{m=1}^M \alpha_m^2 \;\longrightarrow\; \varphi^{-3} = \sqrt5-2 \approx 0.236068,
\qquad
\Meff \;\longrightarrow\; \varphi^{3} = 2+\sqrt5 \approx 4.236068,
\]
independently of $M$. Consequently, for orthonormalized learners with common
variance $\sigma^2$,
\[
\Var\big(\widehat f_{\mathrm{Fib}}\big) \to \varphi^{-3}\sigma^2,
\qquad
\Var\big(\widehat f_{\mathrm{unif}}\big) = \sigma^2/M \to 0 .
\]
\end{theorem}

\begin{proof}
By Lemma~\ref{lem:golden},
$\sum_m\alpha_m^2 \to \sum_{k\ge0}\varphi^{-2(k+2)} = \varphi^{-4}/(1-\varphi^{-2})$.
Using $\varphi^2=\varphi+1$,
\[
\frac{\varphi^{-4}}{1-\varphi^{-2}}
= \frac{\varphi^{-4}\varphi^{2}}{\varphi^{2}-1}
= \frac{\varphi^{-2}}{\varphi} = \varphi^{-3}.
\]
The variance statements follow from orthonormality.
\end{proof}

\begin{table}[t]
\centering
\caption{Fibonacci weights concentrate on a bounded number of learners. The
effective size saturates at $\varphi^3\approx4.236$ and does not grow with $M$.}
\label{tab:effsize}
\begin{tabular}{rrrr}
\toprule
$M$ & $\sum_m\alpha_m^2$ & $1/M$ & $\Meff$\\
\midrule
5   & 0.27778 & 0.20000 & 3.600\\
10  & 0.23938 & 0.10000 & 4.178\\
20  & 0.23609 & 0.05000 & 4.236\\
50  & 0.23607 & 0.02000 & 4.236\\
100 & 0.23607 & 0.01000 & 4.236\\
500 & 0.23607 & 0.00200 & 4.236\\
\bottomrule
\end{tabular}
\end{table}

\begin{corollary}[No asymptotic variance advantage]\label{cor:novar}
In the orthogonalized homoscedastic regime, uniform weighting strictly dominates
Fibonacci weighting in variance for every $M > \varphi^3 \approx 4.24$.
\end{corollary}

Table~\ref{tab:effsize} shows the saturation setting in almost immediately: by
$M=20$ the effective size has reached its limit to three decimals. The reading is
blunt but clarifying. Adding base learners to a Fibonacci ensemble does not make
it more stable. Whatever such a scheme achieves, it does not achieve variance
reduction by aggregation, which is the classical justification for bagging and
random forests. Its case must be made on bias, and Section~\ref{sec:bv} makes it
precisely.

\begin{remark}[Why the golden ratio is still present]
It would be a misreading to conclude that $\varphi$ plays no role. It plays a
sharp one: it \emph{is} the ceiling. The quantity $\varphi^3$ is the exact number
of learners a Fibonacci ensemble can consult, and the proportion that seemed to
promise expressive reach turns out to measure a limit. This is a more interesting
role than the one originally envisaged, and it is exact.
\end{remark}

\section{Bias--Variance Structure}\label{sec:bv}

We now determine precisely when concentrating weight on part of an ordered
dictionary is worthwhile. Let $\{h_m\}_{m=1}^M$ be orthonormal in $L^2(P_X)$,
let $f^\star = \sum_m \theta_m h_m$, and suppose that from $n$ observations we
form uncorrelated unbiased estimates $\widehat\theta_m$ with
$\Var(\widehat\theta_m) = \tau_m^2/n$; this holds for orthogonal series least
squares under homoscedastic noise \citep{Gyorfi2002, Tsybakov2009}. For $w$ on
the simplex write $\widehat f_w = \sum_m w_m\widehat\theta_m h_m$ and
$\mathcal R(w) = \E\|\widehat f_w - f^\star\|_2^2$.

\begin{theorem}[Exact crossover for geometric weighting]\label{thm:crossover}
The integrated risk is
\[
\mathcal R(w) = \sum_{m=1}^M\Big[(w_m-1)^2\theta_m^2 + w_m^2\frac{\tau_m^2}{n}\Big],
\]
and a geometric law $w^{(r)}_m \propto r^m$ satisfies
$\mathcal R(w^{(r)}) < \mathcal R(w^{\mathrm{unif}})$ if and only if
\[
\underbrace{\sum_m\Big[(w^{(r)}_m-1)^2-\big(\tfrac1M-1\big)^2\Big]\theta_m^2}_{\Delta B(r)}
\;<\;
\underbrace{\frac1n\sum_m\Big[\tfrac1{M^2}-\big(w^{(r)}_m\big)^2\Big]\tau_m^2}_{\Delta V(r)} .
\]
In the homoscedastic case $\tau_m^2\equiv\tau^2$,
$\Delta V(r) = \frac{\tau^2}{n}\big(\frac1M - \sum_m (w^{(r)}_m)^2\big) < 0$ for
every $r\neq1$, so a geometric law can win only by strictly reducing bias.
\end{theorem}

\begin{proof}
Orthonormality gives $\widehat f_w - f^\star = \sum_m(w_m\widehat\theta_m-\theta_m)h_m$,
and unbiasedness with uncorrelated errors yields the stated risk. Subtracting the
uniform risk and rearranging gives the condition. For the sign of $\Delta V$,
$\sum_m w_m^2 > 1/M$ for any non-uniform $w$ on the simplex, equivalently
$\Meff(w) < M$ by Definition~\ref{def:eff}.
\end{proof}

Theorem~\ref{thm:crossover} states the scope condition of the whole family in
falsifiable form. Geometric and Fibonacci weighting are bias-reduction devices
for ordered dictionaries. They pay a variance premium that grows with $M$, and
they recover it only when the coefficient energy $\theta_m^2$ is concentrated on
the indices that $r^m$ upweights. When the signal sits at low indices the optimal
ratio satisfies $r<1$; when it sits at high indices, $r>1$.
Section~\ref{sec:exp} exhibits both regimes.

\subsection{When Orthogonal Weighting Is Optimal}

The following result and its corollary delineate the complementary regime.

\begin{theorem}[Optimality of inverse-variance weighting in the redundant regime]
\label{thm:rb}
In the orthonormal setting above, suppose $\theta_m = 0$ for all $m>K$, and let
$\widehat f_{\mathrm{IV}}$ use the oracle weights
\[
w^{\mathrm{IV}}_m =
\begin{cases}
\tau_m^{-2}\big/\sum_{j\le K}\tau_j^{-2}, & m \le K,\\
0, & m>K.
\end{cases}
\]
Then $\mathcal R(w^{\mathrm{IV}}) \le \mathcal R(w)$ for every $w$ supported on
$\{1,\dots,K\}$ summing to one, with equality iff $w = w^{\mathrm{IV}}$; in
particular $\mathcal R(w^{\mathrm{IV}}) \le \mathcal R(w^{\mathrm{Fib}})$.
\end{theorem}

\begin{proof}
Any positive weight outside $\{1,\dots,K\}$ contributes pure variance with no
bias reduction, so an optimal $w$ sets $w_m=0$ for $m>K$. The minimization then
reduces to a strictly convex quadratic program over the simplex, whose unique
minimizer is $w_m \propto \tau_m^{-2}$. Strict convexity gives the equality
condition.
\end{proof}

\begin{corollary}[A regime where Fibonacci can dominate]\label{cor:fibwins}
Suppose instead that $\{h_m\}$ form an overcomplete, moderately correlated
dictionary, that the effective signal is spread over many coordinates so that
truncation induces substantial bias, and that the variances $\tau_m^2$ grow
moderately with $m$. Then a truncated inverse-variance estimator attains low
variance at the cost of noticeable approximation bias, while a geometric or
Fibonacci-weighted estimator built from the full dictionary retains more
high-frequency structure at controlled variance cost, and can achieve strictly
smaller total risk.
\end{corollary}

\begin{remark}[Taxonomy]
Theorem~\ref{thm:rb} and Corollary~\ref{cor:fibwins} together give a simple
picture. In low-dimensional, orthogonal, strongly redundant bases,
inverse-variance weighting on a truncated support is optimal and structured
weighting cannot improve on it. In rich, overcomplete, moderately correlated
dictionaries, structured weighting can reduce approximation bias enough to pay
for its variance premium. Which regime obtains is an empirical question about the
dictionary, not a matter of principle---and Theorem~\ref{thm:crossover} says
exactly which quantity decides it.
\end{remark}

\section{Gram Whitening and Orthogonalized Ensembles}

Where base learners are correlated it is convenient to precondition. Let
$G_{m,m'} = \langle h_m, h_{m'}\rangle$ be invertible and set
$h^{\perp} = G^{-1/2}h$.

\begin{proposition}[Whitening]\label{prop:whiten}
$\langle h^\perp_m, h^\perp_{m'}\rangle = \delta_{mm'}$, and for any $w$,
$\Var\big(\sum_m w_m h^\perp_m\big) = \sum_m w_m^2\sigma_m^2$, so whitening
removes the cross-covariance term from the variance decomposition.
\end{proposition}

Two cautions are worth stating plainly. First, whitening is a change of basis
within $\operatorname{span}\{h_m\}$: it simplifies variance accounting but does
not by itself reduce risk, and $\sum_m w_m h_m^\perp$ estimates a different
functional from $\sum_m w_m h_m$. Second, $G^{-1/2}$ costs $O(M^3)$ and is
ill-conditioned precisely in the overcomplete regime where structured weighting
is most attractive; we therefore use the ridge-stabilized
$(G+\varepsilon\,\overline{\operatorname{tr}}(G)I)^{-1/2}$ with
$\varepsilon=10^{-8}$ throughout, and recommend reporting $\kappa(G)$ alongside
any result that relies on whitening.

\section{Second-Order Recursive Ensemble Flows}\label{sec:flows}

Boosting is a first-order recursion, $F_m = F_{m-1} + \Delta F_{m-1}$. A
structured alternative introduces memory of the preceding two states,
\begin{equation}\label{eq:flow}
F_m = \beta F_{m-1} + \gamma F_{m-2} + \Delta F_{m-1}, \qquad m\ge3,
\end{equation}
with companion matrix
$T = \left(\begin{smallmatrix}\beta&\gamma\\1&0\end{smallmatrix}\right)$ and
characteristic polynomial $\lambda^2-\beta\lambda-\gamma$. The term
$\gamma F_{m-2}$ introduces memory: the system evolves not only from where it is
but from where it has just been.

\subsection{Stability Region}

\begin{proposition}[Schur--Cohn stability region]\label{prop:stab}
The homogeneous part of \eqref{eq:flow} is $L^2$-stable, i.e. $\rho(T)<1$, if and
only if
\[
|\gamma| < 1, \qquad 1-\beta-\gamma > 0, \qquad 1+\beta-\gamma > 0,
\]
equivalently $(\beta,\gamma)$ lies in the open triangle with vertices
$(-2,-1)$, $(2,-1)$, $(0,1)$. For $\beta,\gamma\ge0$ this reduces to
$\beta+\gamma<1$.
\end{proposition}

\begin{proof}
Apply the Schur--Cohn criterion to $\lambda^2+a_1\lambda+a_0$ with $a_1=-\beta$,
$a_0=-\gamma$: all roots lie strictly inside the unit disk iff $|a_0|<1$,
$1+a_1+a_0>0$, $1-a_1+a_0>0$.
\end{proof}

\begin{remark}[What the region excludes]
The inequality $\beta^2+4\gamma<4$ separates real from complex roots and is
\emph{not} sufficient for stability: $(\beta,\gamma)=(1.9,0.01)$ satisfies it yet
gives $\rho(T)=1.905$, and $(1.5,0.4)$ gives $\rho(T)=1.731$. More importantly,
the classical Fibonacci choice $(\beta,\gamma)=(1,1)$ has $\rho(T)=\varphi>1$ and
lies outside the stable region: the unnormalized Fibonacci flow diverges.
\end{remark}

\begin{remark}[Damping the forcing term does not suffice]
It is tempting to stabilize the Fibonacci flow by shrinking the step sizes,
taking $\Delta F_{m-1} = \eta_m h_{m-1}$ with $\eta_m \le C\varphi^{-m}$. This
does not work. Writing $F_T = \sum_k \alpha_{T,k}h_k$, the coefficients obey the
Fibonacci recurrence in $T$, so $\alpha_{T,k} \sim \varphi^{T-1-k}\eta_k$ and
$\sum_k|\alpha_{T,k}| \sim \varphi^{T-1}\sum_k \varphi^{-2k} \to \infty$. The
homogeneous factor $\varphi^T$ multiplies every forcing term regardless of how
fast $\eta$ decays; numerically, $\sum_k|\alpha_{T,k}|$ reaches $1.1\times10^{4}$
at $T=20$ and $3.8\times10^{16}$ at $T=80$. Stability must be obtained by damping
$\beta$ and $\gamma$ themselves.
\end{remark}

\subsection{The Damped Fibonacci Flow}

\begin{definition}[Damped Fibonacci flow]\label{def:damped}
Fix $\delta\in(0,1)$ and set
$\beta_\delta = \delta(1-\varphi^{-2})$, $\gamma_\delta = \delta^2\varphi^{-2}$.
\end{definition}

\begin{proposition}\label{prop:damped}
The flow of Definition~\ref{def:damped} has characteristic roots
$\lambda_+ = \delta$ and $\lambda_- = -\delta\varphi^{-2}$. Hence
$\rho(T)=\delta<1$ and the Schur--Cohn conditions hold for every
$\delta\in(0,1)$, while the mode ratio is preserved at
\[
\frac{\lambda_+}{|\lambda_-|} = \varphi^{2},
\]
exactly as for the classical recursion, whose roots are $\varphi$ and
$-\varphi^{-1}$.
\end{proposition}

\begin{proof}
By construction $\lambda_++\lambda_- = \beta_\delta$ and
$\lambda_+\lambda_- = -\gamma_\delta$, so these are the roots of
$\lambda^2-\beta_\delta\lambda-\gamma_\delta$; stability is immediate from
$\rho(T)=\delta$. For $\delta=0.9$, $(\beta,\gamma)=(0.5562,0.3094)$ with
$1-\beta-\gamma = 0.1344>0$ and $1+\beta-\gamma = 1.2468>0$. Numerically,
$\sum_k|\alpha_{T,k}|$ decays from $1.31$ at $T=5$ to $6\times10^{-4}$ at $T=80$.
\end{proof}

The golden ratio therefore survives in the recursive setting, but as a
\emph{relative} spectral signature rather than as a growth rate: it is the ratio
of the two modes, not their magnitude, that carries the Fibonacci character. This
seems to us the right home for it. A dominant mode of magnitude $\varphi$ is
simply an instability; a mode \emph{ratio} of $\varphi^2$ is a structural
property that can be preserved at any stability level.

\section{Algorithms}

\subsection*{Algorithm 1: Fibonacci-Weighted Ensemble}
\begin{enumerate}
\item Train base learners $h_1,\dots,h_M$ on $\mathcal D_n$, \emph{ordered by
increasing complexity}.
\item Optionally form $G_{m,m'} = \frac1n\sum_i h_m(x_i)h_{m'}(x_i)$ and whiten,
$h^\perp = (G+\varepsilon\,\overline{\operatorname{tr}}(G)I)^{-1/2}h$.
\item Compute $\alpha_m = F_m/\sum_j F_j$.
\item Return $\widehat f(x) = \sum_m \alpha_m h^{(\perp)}_m(x)$.
\end{enumerate}

\noindent
Step 1's ordering requirement is essential rather than incidental; see
Section~\ref{sec:perm}.

\subsection*{Algorithm 2: Damped Fibonacci Recursive Flow}
\begin{enumerate}
\item Choose $\delta\in(0,1)$ and set
$\beta_\delta = \delta(1-\varphi^{-2})$, $\gamma_\delta = \delta^2\varphi^{-2}$.
\item Initialize $F_1, F_2$ with simple regressors.
\item For $m=3,\dots,M$: compute residuals $r_{m-1} = y - F_{m-1}$, fit $h_{m-1}$
to $r_{m-1}$, and update
$F_m = \beta_\delta F_{m-1} + \gamma_\delta F_{m-2} + h_{m-1}$.
\item Return $F_M$.
\end{enumerate}

\section{Computational Experiments}\label{sec:exp}

Theorem~\ref{thm:effsize} shows that Fibonacci weights are asymptotically
geometric with ratio $\varphi$. The decisive empirical question is therefore
whether $r=\varphi$ is a preferred point of the family $w_m\propto r^m$.

\paragraph{Setup.}
Targets $f_{\sin}(x)=\sin(2\pi x)$ on $[0,1]$ and
$f_{\mathrm{sinc}}(x)=\sin(x)/x$ on $[-3,3]$; $n_{\mathrm{train}}=200$,
$n_{\mathrm{test}}=1000$, Gaussian noise $\sigma=0.3$, ridge parameter
$\lambda=10^{-3}n$, $50$ Monte Carlo replications under common random numbers.
Dictionaries: random Fourier feature ridge learners \citep{RahimiRecht2007RFF,
ScholkopfSmola2002} with $D=60$ features and lengthscales log-spaced over
$[0.60,0.03]$, indexed smooth to oscillatory; and polynomial ridge learners of
degree $1,\dots,M$. ISE is computed on a $2000$-point grid, normalized by
interval length. Code and seeds are released with this manuscript.

\subsection{Fibonacci Weighting Is Geometric Weighting at $r=\varphi$}

The maximum absolute deviation between $\alpha^{\mathrm{Fib}}$ and $w^{(\varphi)}$
is $2.2\times10^{-2}$ at $M=5$, $1.9\times10^{-3}$ at $M=10$,
$1.6\times10^{-5}$ at $M=20$, and below machine precision for $M\ge40$; the
effective sizes agree to three decimals for $M\ge20$. Any claim specific to the
Fibonacci sequence is therefore a claim about the number $\varphi$.

\subsection{$\varphi$ Is Not Generally Optimal}

\begin{table}[t]
\centering
\caption{Risk-optimal geometric ratio $r^\star$ (minimizing mean ISE over
$r\in[0.2,5]$) against the Fibonacci value $\varphi\approx1.618$. Final column:
multiplicative penalty from using $\varphi$ in place of $r^\star$.}
\label{tab:rsweep}
\begin{tabular}{llrrrrr}
\toprule
Dictionary & Target & $M$ & $r^\star$ & ISE$(r^\star)$ & ISE$(\varphi)$ & penalty\\
\midrule
RFF        & $\sin$ & 10 & 1.10 & 0.0042 & 0.0062 & $1.47\times$\\
RFF        & $\sin$ & 40 & 1.00 & 0.0038 & 0.0088 & $2.33\times$\\
RFF        & sinc   & 10 & 0.20 & 0.0041 & 0.0849 & $20.97\times$\\
RFF        & sinc   & 40 & 0.30 & 0.0040 & 0.1413 & $35.66\times$\\
Polynomial & $\sin$ & 10 & 2.20 & 0.0608 & 0.0618 & $1.02\times$\\
Polynomial & $\sin$ & 40 & 1.20 & 0.0514 & 0.0515 & $1.00\times$\\
Polynomial & sinc   & 10 & 1.30 & 0.0024 & 0.0024 & $1.03\times$\\
Polynomial & sinc   & 40 & 0.95 & 0.0028 & 0.0047 & $1.67\times$\\
\bottomrule
\end{tabular}
\end{table}

Table~\ref{tab:rsweep} reports $r^\star$ across eight configurations. It ranges
from $0.20$ to $2.20$ and depends on dictionary, target and $M$. In the
polynomial experiments at $M=10$, $\varphi$ falls within $3\%$ of the optimum and
is a defensible default. In the RFF/sinc experiments it is worse by factors of
$21$ to $36$, and the optimum lies at $r<1$: weighting the early, smoother
learners most heavily, the opposite of the Fibonacci prescription. This is
exactly the behaviour Theorem~\ref{thm:crossover} predicts when coefficient
energy sits at low indices.

\begin{figure}[t]\centering
\includegraphics[width=\textwidth]{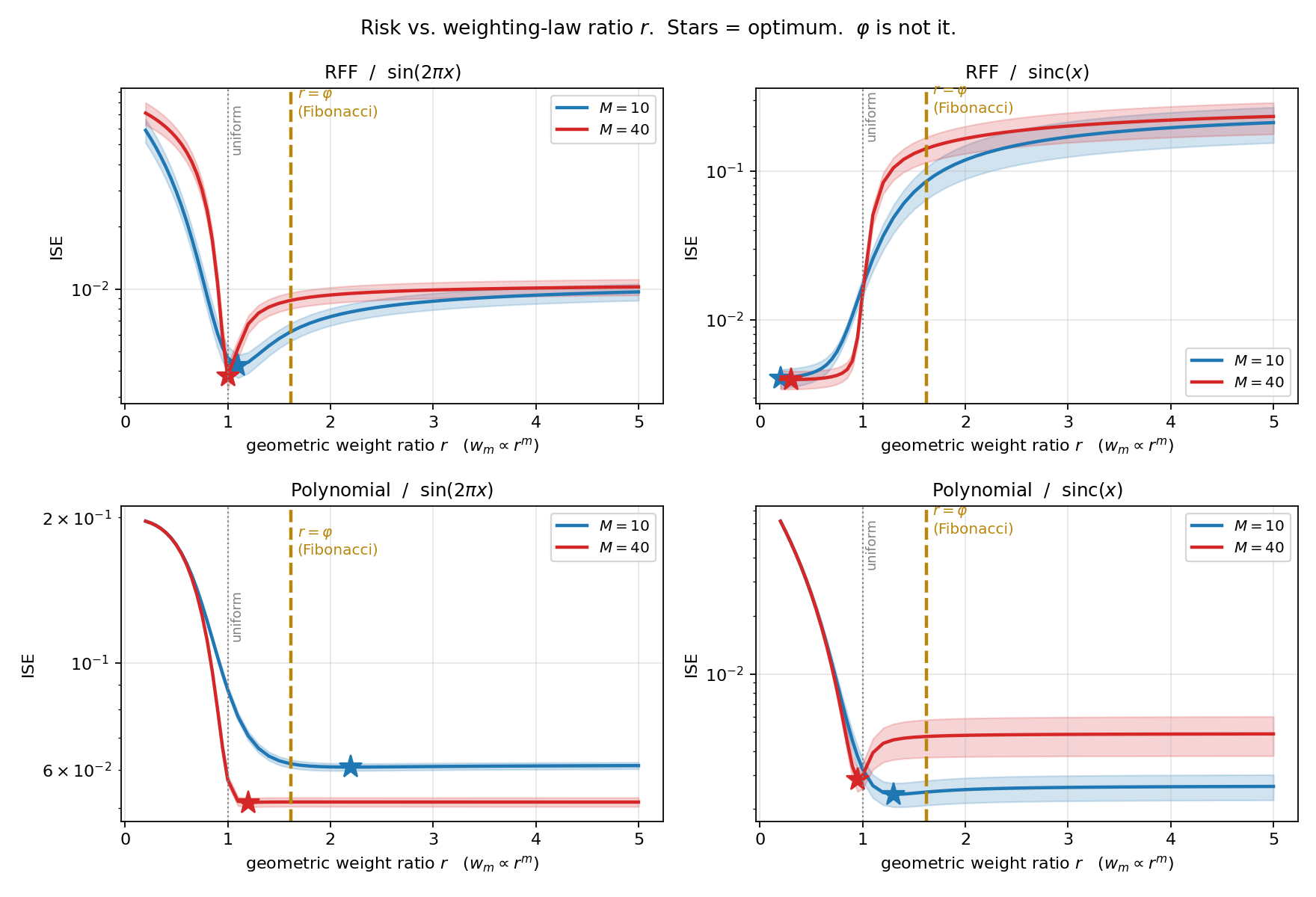}
\caption{Mean ISE against geometric ratio $r$ over $50$ replications (shaded:
$\pm2$ standard errors). Dashed line: $r=\varphi$. Stars: empirical optimum.}
\label{fig:rsweep}
\end{figure}

\subsection{The Choice of Law Matters by an Order of Magnitude}

Figure~\ref{fig:rsweep} shows ISE varying by more than an order of magnitude
across $r$. This is the central positive finding: the aggregation rule, the one
component of the pipeline routinely inherited rather than designed, is worth an
order of magnitude in risk. That no fixed $r$ serves all regimes is precisely why
the law deserves to be estimated rather than assumed.

\subsection{Risk Degrades with $M$, as Theorem~\ref{thm:effsize} Predicts}

\begin{figure}[t]\centering
\includegraphics[width=\textwidth]{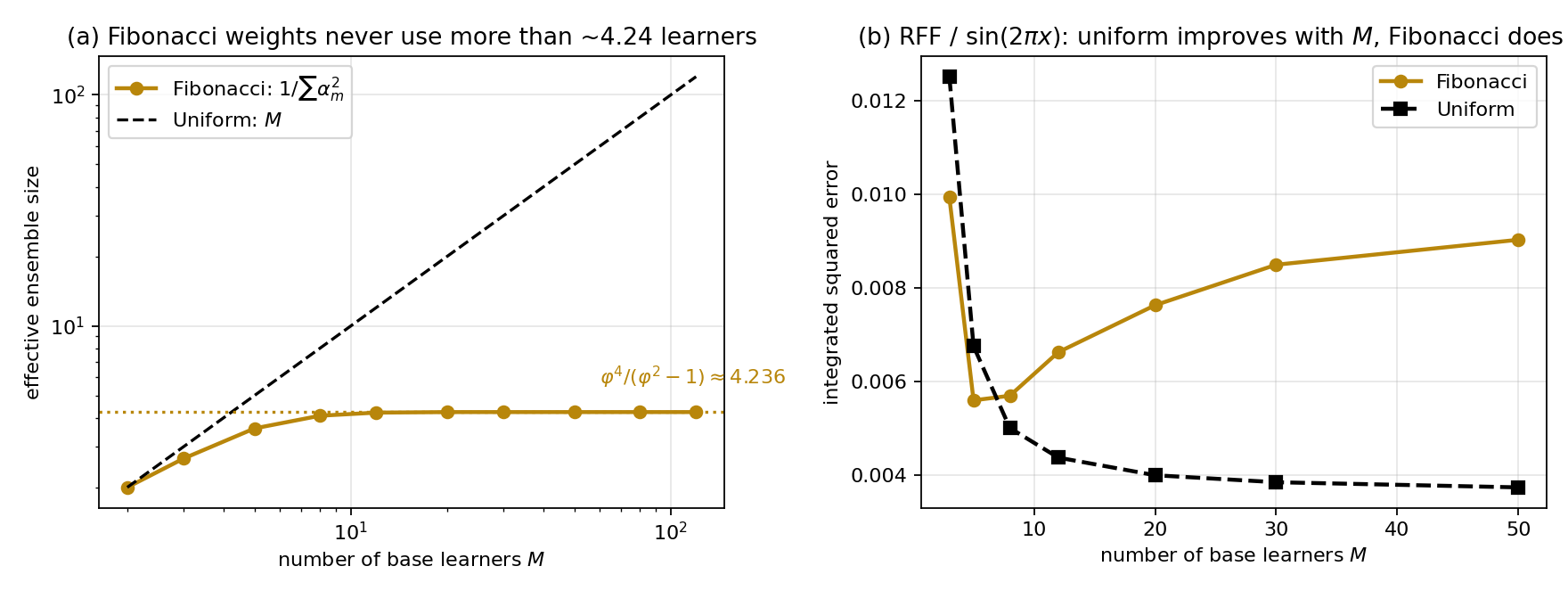}
\caption{(a) Effective ensemble size saturates at $\varphi^3\approx4.236$
regardless of $M$. (b) Consequently uniform averaging improves with $M$ while
Fibonacci weighting does not.}
\label{fig:effsize}
\end{figure}

On RFF with $f_{\sin}$, uniform ISE improves monotonically from $0.0064$ at $M=5$
to $0.0038$ at $M=40$, while Fibonacci ISE \emph{worsens} from $0.0055$ to
$0.0087$. Adding learners helps uniform averaging and hurts Fibonacci weighting
(Figure~\ref{fig:effsize}), a direct empirical signature of the bounded effective
size.

\subsection{Order Dependence}\label{sec:perm}

\begin{figure}[t]\centering
\includegraphics[width=\textwidth]{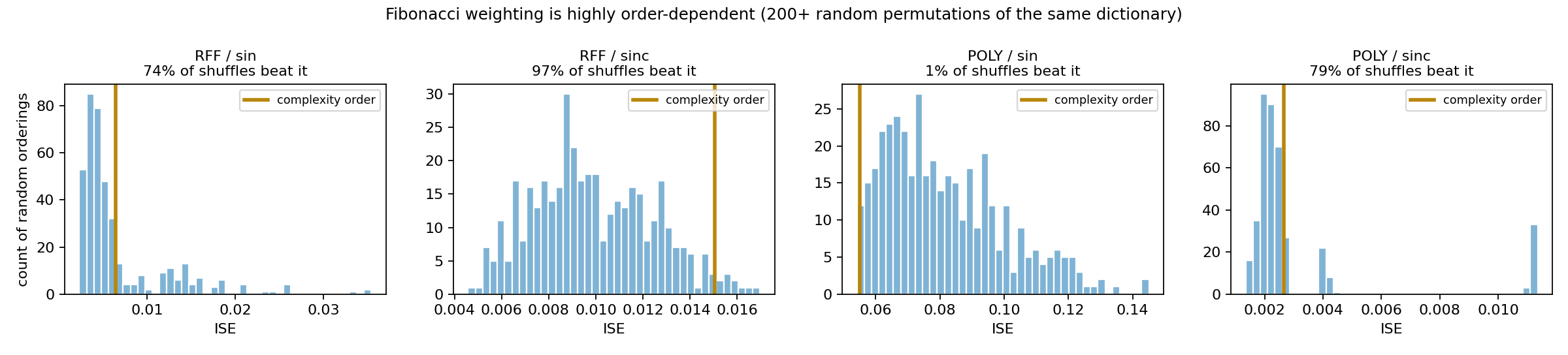}
\caption{ISE under $200$ random permutations of the same dictionary, with the
complexity-sorted ordering marked.}
\label{fig:perm}
\end{figure}

Fibonacci weights place roughly $62\%$ of their mass on the last two learners, so
the index must carry meaning. Permuting the dictionary at random ($200$
permutations, $M=10$) changes ISE by factors of $2.5$ to $15$. On RFF/sinc, $98\%$
of random orderings outperform the complexity-sorted ordering; on RFF/$\sin$,
$71\%$ do; on polynomial/$\sin$, only $1\%$ do (Figure~\ref{fig:perm}).

Two consequences follow. Structured weighting is appropriate only for
dictionaries whose index is informative, and the correct \emph{direction} of the
ordering must be established rather than assumed. For exchangeable dictionaries,
such as bootstrap replicates or independently seeded random features, Fibonacci
weighting reduces to selecting roughly four learners at random---which is a
description of what the method does, not a criticism of it, provided one knows
that is what one is doing.

\section{Comparison with Classical Ordered Shrinkage}\label{sec:classical}

Section~\ref{sec:exp} compared weighting laws against uniform averaging. That is
the natural baseline within the ensemble literature, but it is not the right
benchmark: the question of how influence should decay along an ordered index has
a mature answer in nonparametric estimation, and any proposal of the present kind
should be measured against it. We do so here, and the comparison turns out to be
more informative than we anticipated.

\subsection{Two Constraint Sets}

All estimators considered in this section are \emph{diagonal linear} estimators
in an orthonormal basis,
\begin{equation}\label{eq:diaglin}
\widehat f_w = \sum_{m=1}^M w_m\,\widehat\theta_m\, h_m ,
\qquad
\mathcal R(w) = \sum_{m=1}^M\Big[(w_m-1)^2\theta_m^2 + w_m^2\,v\Big],
\quad v = \sigma^2/n ,
\end{equation}
with the risk identity exact by orthonormality. They differ only in the set over
which $w$ is allowed to range:
\begin{align}
\text{aggregation (simplex)}: &\qquad \textstyle\sum_m w_m = 1,\quad w_m\ge0,
\label{eq:simplex}\\
\text{ordered shrinkage}: &\qquad w_m\in[0,1] \text{ for each } m .
\label{eq:shrink}
\end{align}
Constraint~\eqref{eq:simplex} is the ensemble convention: bagging, stacking, the
SuperLearner, and Fibonacci weighting all combine predictors convexly.
Constraint~\eqref{eq:shrink} is the convention of Pinsker \citep{Pinsker1980},
blockwise Stein shrinkage, and ordered linear smoothers
\citep{Kneip1994Ordered, CavalierTsybakov2002}, which shrink coefficients
individually and impose no budget across the index.

The distinction is easy to overlook because both are ``weights'', but it is
consequential, and separating the two is the main methodological point of this
section.

\subsection{Estimators Compared}

\begin{description}
\item[Uniform.] $w_m = 1/M$.
\item[Fibonacci.] $w_m = F_m/\sum_j F_j$.
\item[Best geometric.] $w_m \propto r^m$ with $r$ minimizing $\mathcal R$ over a
grid; an oracle within the geometric family.
\item[Best simplex.] The exact minimizer of $\mathcal R$ over~\eqref{eq:simplex}.
By Karush--Kuhn--Tucker conditions,
$w_m = \max\{0,(\theta_m^2+\mu)/(\theta_m^2+v)\}$ with $\mu$ fixed by
$\sum_m w_m = 1$; see Appendix~\ref{app:simplex}. This is the best any
simplex-constrained scheme can do.
\item[Pinsker.] $w_m = (1-a_m/\mu)_+$ on the Sobolev ellipsoid $a_m = m^{\beta}$,
$\beta=2$, with $\mu$ calibrated from the data \citep{Pinsker1980}.
\item[Blockwise Stein.] James--Stein shrinkage applied on weakly geometric blocks.
\item[Oracle shrinkage.] $w_m = \theta_m^2/(\theta_m^2+v)$, the ideal diagonal
linear estimator.
\end{description}

We take $M=40$, $n=200$, $\sigma=0.3$, $\theta_m \propto m^{-a}$ normalized to
$\|\theta\|=2$, and average over $200$ replications.

\begin{table}[t]
\centering
\caption{Sequence model, $M=40$, $n=200$, $\sigma=0.3$, $200$ replications.
Integrated risk. The horizontal rule separates simplex-constrained
estimators~\eqref{eq:simplex} from ordered shrinkage~\eqref{eq:shrink}.}
\label{tab:classical}
\small
\begin{tabular}{lrrrr}
\toprule
& \multicolumn{4}{c}{signal decay exponent $a$}\\
\cmidrule(lr){2-5}
Estimator & $a=1.0$ & $a=1.5$ & $a=2.0$ & $a=3.0$\\
\midrule
Uniform            & 3.8025 & 3.8025 & 3.8025 & 3.8025\\
Fibonacci          & 3.9971 & 4.0000 & 4.0001 & 4.0001\\
Best geometric ($r^\star$) & 1.4266 & 0.6313 & 0.2922 & 0.0727\\
Best simplex       & 1.4081 & 0.6257 & 0.2910 & 0.0677\\
\midrule
Pinsker            & 0.0165 & 0.0106 & 0.0088 & 0.0087\\
Blockwise Stein    & 0.0166 & 0.0106 & 0.0057 & 0.0030\\
Oracle shrinkage   & 0.0165 & 0.0094 & 0.0045 & 0.0019\\
\midrule
$r^\star$          & 0.18 & 0.10 & 0.05 & 0.05\\
\bottomrule
\end{tabular}
\end{table}

\subsection{Three Findings}

\paragraph{The geometric family is nearly optimal within the simplex.}
Across $a\in[0.6,3.2]$, the ratio of best-geometric risk to best-simplex risk
lies between $1.004$ and $1.119$. The two-parameter-free geometric shape is thus
an excellent parametrization of simplex-constrained aggregation: essentially
nothing is lost by restricting attention to laws of the form $w_m\propto r^m$.
This is a genuine vindication of the family that motivates this paper, and it is
worth stating clearly before the less comfortable findings.

\paragraph{The golden ratio is the wrong member of it.}
The risk-minimizing ratio satisfies $r^\star\in[0.05,0.27]$ throughout this
range---far below $1$, and hence far below $\varphi$. Fibonacci weighting is
between $1.5$ and $70$ times worse than the best geometric law, with the gap
widening as the signal concentrates on low indices. This is precisely the
behaviour Theorem~\ref{thm:crossover} predicts: when coefficient energy sits at
small $m$, the optimal law upweights early learners, and $\varphi>1$ upweights
exactly the wrong end.

\paragraph{The simplex constraint costs more than the choice of law within it.}
The best simplex-constrained estimator is between $6$ and $146$ times worse than
Pinsker. No choice of weighting law can close that gap, because the gap is a
property of the constraint set, not of the law. Under~\eqref{eq:simplex} the
weights must average $1/M$, so on coordinates carrying real signal
$w_m\widehat\theta_m$ underestimates $\theta_m$ by a factor of order $M$, and the
bias term $(w_m-1)^2\theta_m^2$ dominates. Ordered shrinkage escapes this by
spending no budget at all.

\begin{figure}[t]\centering
\includegraphics[width=\textwidth]{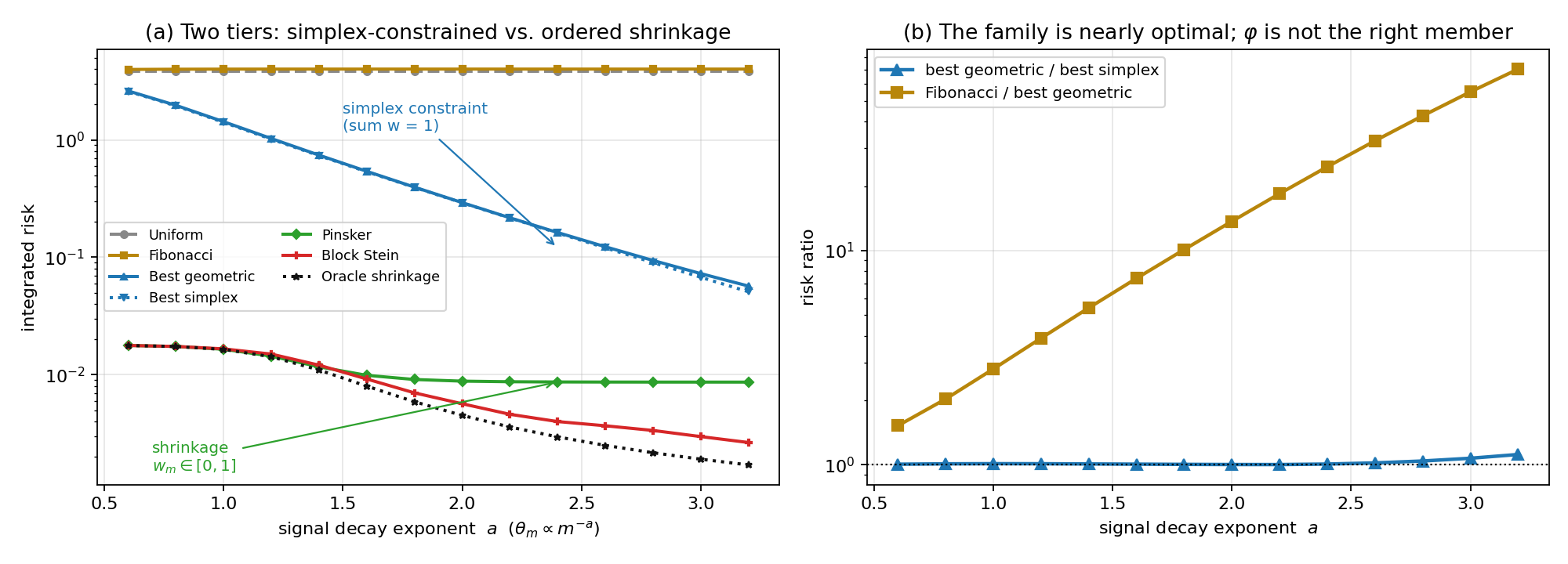}
\caption{(a) Risk against signal decay for the seven estimators. The two
constraint sets separate into distinct tiers roughly two orders of magnitude
apart; Uniform and Fibonacci coincide at the top. (b) Within the simplex the
geometric family is nearly optimal (lower curve, close to $1$), but $\varphi$ is
not the right member of it (upper curve).}
\label{fig:classical}
\end{figure}

\begin{remark}[What this means for ensemble weighting]
The finding is not that structured weighting is useless. It is that the ensemble
convention of convex aggregation---inherited from bagging, and adopted without
comment by most of the aggregation literature including our own earlier
treatment---is itself the binding constraint whenever the base learners are
coefficient-like rather than predictor-like. When each $h_m$ is a full predictor
of $f$, as in bagging or stacking, the simplex is natural and the results of
Section~\ref{sec:exp} apply. When the dictionary is a basis expansion, the
simplex is an artefact, and one should be shrinking rather than averaging. Both
situations arise in practice, and they call for different estimators. We regard
making this distinction explicit as more useful than any particular choice of
$r$.
\end{remark}

\section{Real Data}\label{sec:real}

The experiments above are synthetic by design, since exact risk is available only
when $f^\star$ is known. We now examine three real series with genuine ordered
spectral structure. No ground truth exists, so we report held-out predictive mean
squared error over $100$ random $70/30$ splits, in standardized units, with
$M=20$ random Fourier feature learners ordered from smooth to oscillatory. The
geometric ratio $r$ is selected by an inner validation split of the training data,
so that no test information is used in choosing it.

\begin{description}
\item[Nile.] Annual flow volume of the Nile at Aswan, $1871$--$1970$; $n=100$.
\item[Sunspots.] Annual sunspot activity, $1700$--$2008$; $n=309$.
\item[El Ni\~no.] Monthly Pacific sea-surface temperature, $1950$--$2010$; $n=732$.
\end{description}

\begin{table}[t]
\centering
\caption{Held-out predictive MSE in standardized units ($100$ random $70/30$
splits, $M=20$, standard errors shown). $\bar r$ is the mean cross-validated
geometric ratio. Best entry per row in bold.}
\label{tab:real}
\small
\begin{tabular}{lrrcccccc}
\toprule
Series & $n$ & $\bar r$ & Uniform & Fibonacci & Geometric $r$ & Pinsker & Block Stein & Best single\\
\midrule
Nile     & 100 & 1.82 & \textbf{0.655} & 0.817 & 0.801 & 0.952 & 0.979 & 0.847\\
         &     &      & \small$\pm.017$ & \small$\pm.023$ & \small$\pm.024$ & \small$\pm.027$ & \small$\pm.028$ & \small$\pm.027$\\
Sunspots & 309 & 2.64 & 0.842 & 0.725 & 0.713 & \textbf{0.521} & 0.530 & 0.669\\
         &     &      & \small$\pm.012$ & \small$\pm.013$ & \small$\pm.014$ & \small$\pm.017$ & \small$\pm.017$ & \small$\pm.013$\\
El Ni\~no & 732 & 2.76 & 0.980 & 0.944 & 0.935 & 0.945 & 0.939 & \textbf{0.917}\\
         &     &      & \small$\pm.005$ & \small$\pm.007$ & \small$\pm.010$ & \small$\pm.019$ & \small$\pm.019$ & \small$\pm.016$\\
\bottomrule
\end{tabular}
\end{table}

The three series behave differently, and the differences are interpretable.

\paragraph{Sunspots: ordering informative, shrinkage wins.}
The sunspot series has pronounced periodicity near eleven years, so the
frequency-ordered dictionary is highly informative. Here the pattern of
Section~\ref{sec:classical} reproduces on real data: Pinsker and blockwise Stein
lead at $0.521$ and $0.530$, structured simplex weighting follows at $0.713$ and
$0.725$, and uniform averaging is worst at $0.842$. Structured weighting
substantially improves on uniform---a $15\%$ reduction---but classical shrinkage
improves on structured weighting by considerably more.

\paragraph{Nile: ordering uninformative, uniform wins.}
The Nile record is dominated by a well-documented level shift near $1899$, and is
otherwise close to white around its mean. There is little smooth spectral
structure for a frequency ordering to exploit, and every method that concentrates
weight does worse than plain averaging: uniform attains $0.655$ against $0.801$
for the cross-validated geometric law and $0.952$ for Pinsker. This is a useful
negative control. It confirms in real data what Section~\ref{sec:perm} showed by
permutation: structured weighting requires an informative index, and supplies no
benefit---indeed a penalty---when the index carries no signal.

\paragraph{El Ni\~no: weak predictability throughout.}
Predicting monthly sea-surface temperature from the time index alone is close to
hopeless, and every method sits near $0.92$--$0.98$ in standardized units, that
is, near the variance of the series. Differences are small relative to their
standard errors and we draw no conclusion beyond the observation that no method
is harmed.

\paragraph{On the cross-validated ratio.}
The selected $\bar r$ is $1.82$, $2.64$ and $2.76$ on the three series: greater
than one in each case, and greater than $\varphi\approx1.618$ in each case. On
these data the golden ratio is not the value a practitioner would choose, and
choosing $r$ by cross-validation costs one inner split and is uniformly
preferable to fixing it in advance.

\begin{remark}
Fibonacci weighting is never the best method on any of the three series. It is
also never catastrophic: it beats uniform on two of three and is within $13\%$ of
the cross-validated geometric law throughout. As a tuning-free default it is
defensible; as an optimum it is not supported by these data.
\end{remark}

\section{Discussion and Future Work}

The natural next step is to estimate the weighting law from data. Within the
geometric family this means selecting $r$, for which cross-validation is the
obvious first candidate and the Pinsker \citep{Pinsker1980} and ordered-smoother
\citep{Kneip1994Ordered} constructions the natural benchmarks; a systematic
comparison against these, rather than against uniform averaging, is the
experiment we would most like to see next. A more ambitious formulation places a
prior over laws and treats the index profile as the object of inference,
connecting the framework to Bayesian model averaging \citep{Hoeting1999BMA}.

Extension beyond one dimension raises the question of what \emph{ordering} means
for a high-dimensional dictionary. Section~\ref{sec:perm} suggests that this,
rather than the choice of law, may be the binding constraint in practice.
Tree-based dictionaries, excluded here because their variance profiles are
non-monotone and resist analytic whitening, are an important target: a structural
prior over stagewise contributions in boosted trees is a direct application. The
recursive flows of Section~\ref{sec:flows} are developed in
\citet{FokoueIII}, and the general weighting theory of which Fibonacci weighting
is one instance in \citet{FokoueII}.

Finally, the tuning-free character of geometric laws is attractive in low-compute
and small-sample settings, and evaluation on applied problems with naturally
ordered spectral dictionaries---climate and hydrological forecasting among
them---seems to us the most useful immediate direction.

\section{Conclusion}

We set out to ask whether the Fibonacci sequence supplies a principled
aggregation law for ensemble learning. The answer that emerges is more precise
than a yes and more interesting than a no.

Normalized Fibonacci weights are geometric weights at $r=\varphi$, with an
effective ensemble size fixed at $\varphi^3 = 2+\sqrt5$ regardless of how many
learners are available. They cannot reduce variance in the manner of bagging, and
any advantage must come from bias reduction on a dictionary whose index carries
meaning; Theorem~\ref{thm:crossover} says exactly when that trade is favourable.
The classical Fibonacci recursion is unstable as an ensemble flow, but a damped
variant preserves its golden mode ratio inside the stable region. Empirically the
weighting law moves integrated squared error by more than an order of magnitude,
the optimal geometric ratio varies from $0.20$ to $2.20$ across settings, and
$\varphi$ is nowhere the minimizer.

What survives, and what we commend to the community, is the framing. Aggregation
is a distributional operator on the learner index. It is the least designed
component of the modern ensemble pipeline, it is worth an order of magnitude in
risk, and estimating it from data is a well-posed and consequential problem.

\subsection*{A Contemplative Coda}

This work began from a hunch that the sequence organizing so much of the growth
we observe in the natural world might also organize the way many imperfect
learners are asked to speak with one voice. The mathematics declined the
invitation. The golden ratio is not a privileged weighting law, and the sequence
that suggested it is indistinguishable, past $M=20$, from a plain geometric law.

Yet $\varphi$ did not leave empty-handed. It returned as $\varphi^3$: the exact
and immovable ceiling on how many learners such a scheme can ever consult. The
proportion that seemed to promise expressive reach turns out to measure a limit.
There is an old lesson in that---that a pattern glimpsed in nature is an
excellent reason to ask a question and a poor reason to trust an answer---and it
seems fitting to record it in a paper that set out to argue the opposite.

Ensembles remain, as they were, a record of many partial perspectives woven
together. What this study adds is that the weaving itself has a shape, that the
shape matters more than had been supposed, and that choosing it well is work
still to be done.

\subsection*{Reproducibility}

All experiments use synthetic data. Code, seeds and configuration files
accompany this manuscript.

\appendix
\section{Proofs}\label{app:proofs}

\subsection{Proof of Lemma~\ref{lem:golden}}

Binet's formula gives $F_m = (\varphi^m - \psi^m)/\sqrt5$ with
$\psi = (1-\sqrt5)/2 = -\varphi^{-1}$, so $|\psi|<1$ and
$F_m = \varphi^m/\sqrt5 + O(\varphi^{-m})$. The partial sums satisfy the classical
identity $\sum_{j=1}^M F_j = F_{M+2}-1$, proved by induction: the case $M=1$ reads
$F_1 = 1 = F_3-1 = 2-1$, and if it holds for $M$ then
$\sum_{j\le M+1}F_j = F_{M+2}-1+F_{M+1} = F_{M+3}-1$. Hence for fixed $k$,
\[
\alpha_{M-k}
= \frac{F_{M-k}}{F_{M+2}-1}
= \frac{\varphi^{M-k}/\sqrt5 + O(\varphi^{-M})}{\varphi^{M+2}/\sqrt5 + O(\varphi^{-M})}
\longrightarrow \varphi^{-(k+2)} . \qquad\square
\]

\subsection{Proof of Theorem~\ref{thm:effsize}}

By Lemma~\ref{lem:golden} and dominated convergence (the summands are bounded by
the summable envelope $\varphi^{-2(k+2)}$),
\[
\sum_{m=1}^M \alpha_m^2 \longrightarrow \sum_{k\ge0}\varphi^{-2(k+2)}
= \frac{\varphi^{-4}}{1-\varphi^{-2}} .
\]
Using the defining relation $\varphi^2 = \varphi+1$, equivalently
$\varphi^2-1 = \varphi$,
\[
\frac{\varphi^{-4}}{1-\varphi^{-2}}
= \frac{\varphi^{-4}\,\varphi^{2}}{\varphi^{2}-1}
= \frac{\varphi^{-2}}{\varphi}
= \varphi^{-3}.
\]
Numerically $\varphi^{-3} = \sqrt5-2 = 0.2360679\ldots$, so
$\Meff \to \varphi^3 = 2+\sqrt5 = 4.2360679\ldots$ For orthonormalized learners
with common variance $\sigma^2$ the ensemble variance is
$\sum_m \alpha_m^2\sigma^2$, giving the stated limits; the uniform case is
immediate. $\square$

\subsection{Proof of Corollary~\ref{cor:novar}}

Uniform weights give variance $\sigma^2/M$ and Fibonacci weights
$\sigma^2\sum_m\alpha_m^2$. Since $\sum_m\alpha_m^2$ is decreasing in $M$ with
limit $\varphi^{-3}$ and $\Meff(\alpha)<M$ for all $M>\varphi^3$ by
Theorem~\ref{thm:effsize} together with the monotonicity evident in
Table~\ref{tab:effsize}, the claim follows. $\square$

\subsection{Proof of Theorem~\ref{thm:crossover}}

Orthonormality gives
$\widehat f_w - f^\star = \sum_m (w_m\widehat\theta_m - \theta_m)h_m$ and hence
$\mathcal R(w) = \sum_m \E[(w_m\widehat\theta_m-\theta_m)^2]$. Writing
$w_m\widehat\theta_m - \theta_m = w_m(\widehat\theta_m-\theta_m) + (w_m-1)\theta_m$
and using $\E[\widehat\theta_m] = \theta_m$,
$\Var(\widehat\theta_m) = \tau_m^2/n$, the cross term vanishes and
\[
\mathcal R(w) = \sum_m\Big[(w_m-1)^2\theta_m^2 + w_m^2\tau_m^2/n\Big].
\]
Subtracting $\mathcal R(w^{\mathrm{unif}})$ and grouping the bias and variance
contributions gives the stated equivalence. For the sign of $\Delta V$ under
$\tau_m^2\equiv\tau^2$, Cauchy--Schwarz gives
$1 = (\sum_m w_m)^2 \le M\sum_m w_m^2$, so $\sum_m w_m^2 \ge 1/M$ with equality
iff $w$ is uniform; hence $\Delta V(r) < 0$ for $r\neq1$. $\square$

\subsection{The Best Simplex-Constrained Estimator}\label{app:simplex}

Minimize $\mathcal R(w) = \sum_m[(w_m-1)^2\theta_m^2 + w_m^2 v]$ subject to
$\sum_m w_m = 1$ and $w\ge0$. The objective is strictly convex and the feasible
set compact and convex, so the minimizer is unique. The Lagrangian is
$\mathcal R(w) - 2\mu(\sum_m w_m - 1) - \sum_m \nu_m w_m$ with $\nu_m\ge0$, and
stationarity gives
\[
2(w_m-1)\theta_m^2 + 2w_m v - 2\mu - \nu_m = 0 .
\]
For coordinates with $w_m>0$ complementary slackness forces $\nu_m=0$, whence
$w_m = (\theta_m^2+\mu)/(\theta_m^2+v)$; coordinates with $\theta_m^2+\mu<0$ are
clipped to zero. Therefore
\[
w_m(\mu) = \max\Big\{0,\ \frac{\theta_m^2+\mu}{\theta_m^2+v}\Big\},
\]
and $\mu\mapsto\sum_m w_m(\mu)$ is continuous and nondecreasing, strictly
increasing wherever positive, so the constraint $\sum_m w_m = 1$ determines $\mu$
uniquely. It is computed by bisection in the implementation. $\square$

\subsection{Proof of Proposition~\ref{prop:stab}}

The Schur--Cohn criterion for the monic real quadratic
$p(\lambda) = \lambda^2 + a_1\lambda + a_0$ states that both roots lie strictly
inside the unit disk if and only if $|a_0|<1$, $p(1)>0$ and $p(-1)>0$. With
$a_1 = -\beta$ and $a_0 = -\gamma$ these read $|\gamma|<1$,
$1-\beta-\gamma>0$ and $1+\beta-\gamma>0$. The three inequalities define the
interior of the triangle with vertices $(-2,-1)$, $(2,-1)$, $(0,1)$ in the
$(\beta,\gamma)$ plane. For $\beta,\gamma\ge0$ the third is implied by the second
and the region reduces to $\beta+\gamma<1$. $\square$

\subsection{Proof of Proposition~\ref{prop:damped}}

With $\lambda_+ = \delta$ and $\lambda_- = -\delta\varphi^{-2}$ we have
$\lambda_++\lambda_- = \delta(1-\varphi^{-2}) = \beta_\delta$ and
$-\lambda_+\lambda_- = \delta^2\varphi^{-2} = \gamma_\delta$, so $\lambda_\pm$ are
precisely the roots of $\lambda^2-\beta_\delta\lambda-\gamma_\delta$. Then
$\rho(T) = \max(\delta, \delta\varphi^{-2}) = \delta < 1$ since $\varphi>1$. The
mode ratio is $\lambda_+/|\lambda_-| = \varphi^2$, independent of $\delta$, and
coincides with that of the classical recursion, whose roots $\varphi$ and
$-\varphi^{-1}$ satisfy $\varphi/\varphi^{-1} = \varphi^2$. $\square$

\subsection{Divergence of the Undamped Flow Under Step-Size Decay}
\label{app:divergence}

Consider $F_t = \beta F_{t-1} + \gamma F_{t-2} + \eta_{t-1}h_{t-1}$ with
$(\beta,\gamma) = (1,1)$. Expanding recursively, $F_T = \sum_{k}\alpha_{T,k}h_k$
where the coefficients obey $\alpha_{T+1,k} = \alpha_{T,k} + \alpha_{T-1,k}$ for
$k \le T-2$, with $\alpha_{k+1,k} = \eta_k$. For fixed $k$ the sequence
$T\mapsto\alpha_{T,k}$ therefore satisfies the Fibonacci recursion, so
$\alpha_{T,k} = \Theta(\varphi^{T-1-k}\eta_k)$.

Taking $\eta_k \le C\varphi^{-k}$, as proposed by golden-ratio step-size
schedules, gives $\alpha_{T,k} = \Theta(\varphi^{T-1-2k})$ and hence, with
$\|h_k\|\le B$,
\[
\|F_T\| \;\le\; \sum_{k=0}^{T-1}|\alpha_{T,k}|\,\|h_k\|
\;=\; \Theta\!\Big(B\,\varphi^{T-1}\sum_{k\ge0}\varphi^{-2k}\Big)
\;=\; \Theta\!\big(B\,\varphi^{T}\big),
\]
which diverges. The obstruction is structural: the homogeneous factor
$\varphi^{T-1-k}$ multiplies every forcing term, so no decay rate on $(\eta_k)$
removes the $\varphi^{T}$ growth. Numerically, with $B=1$ and
$\eta_k = \varphi^{-k}$ one finds $\sum_k|\alpha_{T,k}| = 1.09\times10^{4}$ at
$T=20$ and $3.79\times10^{16}$ at $T=80$; taking $\eta_k = \varphi^{-2k}$ gives
$8.86\times10^{3}$ and $3.07\times10^{16}$ respectively. Stability must therefore
be obtained by damping $\beta$ and $\gamma$, as in
Definition~\ref{def:damped}, under which the same quantity decays from $1.31$ at
$T=5$ to $6\times10^{-4}$ at $T=80$. $\square$

\bibliographystyle{plainnat}
\bibliography{refs}
\end{document}